\documentclass{article}

\usepackage{PRIMEarxiv}

\usepackage[utf8]{inputenc} % allow utf-8 input
\usepackage[T1]{fontenc}    % use 8-bit T1 fonts
\usepackage{hyperref}       % hyperlinks
\usepackage{url}            % simple URL typesetting
\usepackage{booktabs}       % professional-quality tables
\usepackage{amsfonts}       % blackboard math symbols
\usepackage{nicefrac}       % compact symbols for 1/2, etc.
\usepackage{microtype}      % microtypography
\usepackage{lipsum}
\usepackage{fancyhdr}       % header
\usepackage{graphicx}       % graphics
\graphicspath{{media/}}     % organize your images and other figures under media/ folder

\usepackage{xcolor}
\usepackage{amsmath,amssymb,amsfonts}
\usepackage{graphicx}
\usepackage{algorithm}
\usepackage{colortbl}
\usepackage{booktabs}
\usepackage{algpseudocode}
\usepackage{comment}

\title{SyncMapV2: Robust and Adaptive Unsupervised Segmentation}

\author{
  Heng~Zhang, Zikang~Wan, and~Danilo~Vasconcellos~Vargas \\
  Department of Information Science and Technology \\
  Kyushu University \\
  Fukuoka, Japan\\
  \texttt{wan.zikang.961@s.kyushu-u.ac.jp} \\
}

\begin{document}
\maketitle

\begin{abstract}
Human vision excels at segmenting visual cues without the need for explicit training, and it remains remarkably robust even as noise severity increases. In contrast, existing AI algorithms struggle to maintain accuracy under similar conditions.
Here, we present SyncMapV2, the first to solve unsupervised segmentation with state-of-the-art robustness.
SyncMapV2 exhibits a minimal drop in mIoU, only 0.01\%, under digital corruption, compared to a 23.8\% drop observed in SOTA methods.
This superior performance extends across various types of corruption: noise (7.3\% vs. 37.7\%), weather (7.5\% vs. 33.8\%), and blur (7.0\% vs. 29.5\%). Notably, SyncMapV2 accomplishes this without any robust training, supervision, or loss functions.
It is based on a learning paradigm that uses self-organizing dynamical equations combined with concepts from random networks.
Moreover, unlike conventional methods that require re-initialization for each new input, SyncMapV2 adapts online, mimicking the continuous adaptability of human vision.
Thus, we go beyond the accurate and robust results, and present the first algorithm that can do all the above online, adapting to input rather than re-initializing. 
In adaptability tests, SyncMapV2 demonstrates near-zero performance degradation, which motivates and fosters a new generation of robust and adaptive intelligence in the near future.
\end{abstract}

\section{Introduction}
The human vision system exhibits a level of robustness that current computer vision systems have yet to achieve \cite{azulay2019deep,recht2018cifar}. Not only are humans not fooled by adversarial attacks, but they also display resilience against diverse forms of visual corruption \cite{cea2014intriguing,goodfellow2014explaining,nguyen2015deep}.
% such as noise, adverse weather conditions, and digital distortions \cite{hendrycks2019benchmarking}.
Numerous defensive and detection mechanisms have been proposed to mitigate robustness issues in deep networks. However, as noted by \cite{kotyan2022adversarial}, no current solutions consistently negate adversarial attacks. Defensive systems are extensively proved to have limitations \cite{buckman2018thermometer,uesato2018adversarial}. Attempts to enhance robustness through robust training \cite{madry2017towards} and detection \cite{xu2017feature} have also been made. These methods, as reported, have not provided a consistent solution to the robustness challenge \cite{carlini2017adversarial}.
Moreover, robustness in vision often coincides with adaptability, a critical feature that allows systems to operate effectively in dynamic environments. For instance, the human visual system not only demonstrates robustness to visual corruptions but also excels in rapidly and adaptively detecting and integrating patterns over time. This capability allows humans to effortlessly classify and respond to visual stimuli in an ongoing, real-time manner, without the need for frequent recalibration.

\begin{figure}[ht]
\centering
\includegraphics[width=0.75\columnwidth]{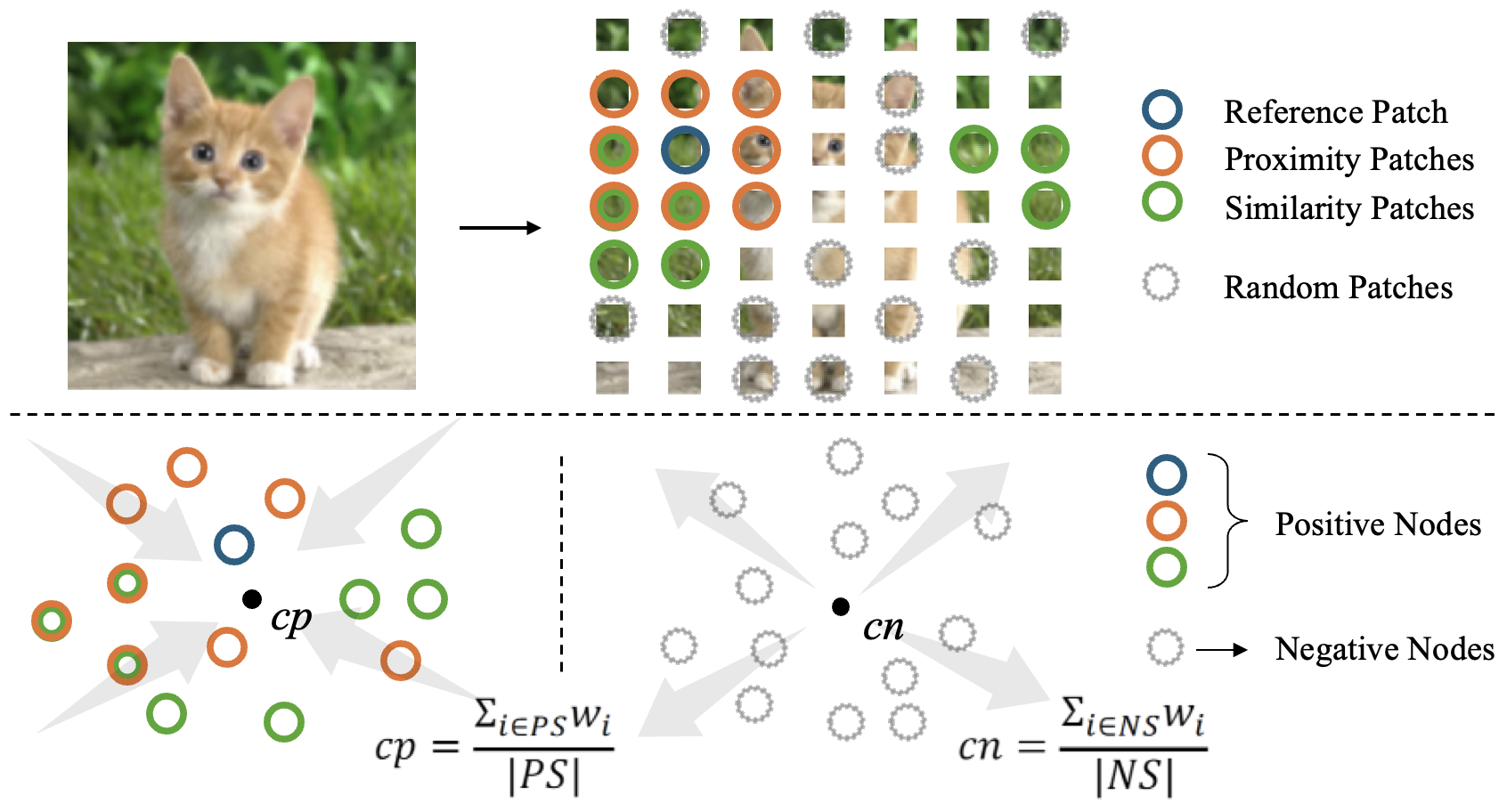} 
% \caption{The comparison between DFC and SyncMapV2 on clean and corrupted images.}
\caption{Illustration of SyncMapV2's dynamics. Note that the random patches are selected apart from positive patches. Patches are then self-organized in SyncMapV2's map space.}
\label{fig:figure1}
\end{figure}

\textcolor{black}{While much of the existing research on robustness in deep learning for vision has centered on supervised learning models, the robustness of unsupervised image segmentation remains underexplored. The original SyncMap \cite{vargas2021syncmap} demonstrated the potential of dynamical systems to learn adaptively and robustly, but it was limited to binary sequences and had never been applied to images or unsupervised segmentation tasks. Beyond these limitations, no algorithm using sequential input is known to work well for segmenting images due to the inherent loss of structural information. }

\textcolor{black}{In contrast, SyncMapV2 is, perhaps, the first method to tackle these challenges by employing a sequential encoding that preserves image structure. By leveraging random networks and dynamical systems to build a sequence based on priorities that maintain intrinsic spatial relationships, SyncMapV2 achieves breakthrough-level robustness—handling various noise corruptions without any specialized robust training or supervision. This advancement pushes the boundaries of robust AI to a significant extent by delivering the brain's ability to integrate spatial information in an online manner and adapt to changing environments.
}

\section{Related Work}
\subsection{Image Segmentation}
Image segmentation involves partitioning an image into distinct regions representing meaningful components like objects or textures. Semantic segmentation, a specialized form, classifies each pixel into predefined semantic classes, typically requiring large labeled datasets and supervised training. 
\textcolor{black}{Although semantic segmentation can be unsupervised, it still requires self-supervised training \cite{niu2024unsupervised, hamilton2022unsupervised}.}
% Semantic segmentation can be unsupervised, which also
% requires self-supervised training \cite{niu2024unsupervised, hamilton2022unsupervised}.
In contrast, our focus is on unsupervised image segmentation, a task that groups pixels based on intrinsic properties such as color or texture without relying on labeled data or pre-training. Unlike unsupervised semantic segmentation, methods in this area operate purely through model inference.

Early approaches to unsupervised image segmentation include k-means \cite{1000236} and graph-based methods \cite{felzenszwalb2004efficient}. Recent advancements feature deep learning-based techniques like W-Net \cite{xia2017w}, Backprops-superpixel (Backprops) \cite{Kanezaki2018Unsupervised}, Invariant Information Clustering (IIC) \cite{Invariant2019iic}, Differentiable Feature Clustering (DFC) \cite{kim2020unsupervised}, and the latest Pixel-level network \cite{hoang2024pixel}. While these methods have set new benchmarks, they predominantly rely on loss functions. In this work, we reveal the inherent drawbacks of such reliance, particularly in terms of robustness and adaptability.

\subsection{Cognition in Representational Spaces}
Recently, neuroscientists suggests cognition as the result of transformations between or movement within representational spaces that are implemented by neural populations \cite{barack2021two}. 
This perspective contrasts with the conventional view that attributes cognitive processes to neurons and their interconnections within a network (e.g., weights' connections, activation functions).
Instead, it relegates the identity of and connections between specific neurons; in an extreme form, it avoids single cell details altogether \cite{trautmann2019accurate}.
This has been proved in a simple memory task in which representational spaces provide explanatory resources unavailable to the conventional view \cite{rigotti2013importance}.
Another example is a type of random networks called reservoir computing, which is a non-linear dynamical system that maps low-dimensional inputs into a high-dimensional neural space, enabling simple readout layers to generate adequate responses for various applications \cite{zhang2023survey}.
In this work, we apply a random network as a feature generator, producing high-dimensional responses that SyncMapV2 processes to make final decisions by its self-organizing characteristic.

\begin{figure*}[ht]
\centering
\includegraphics[width=\textwidth]{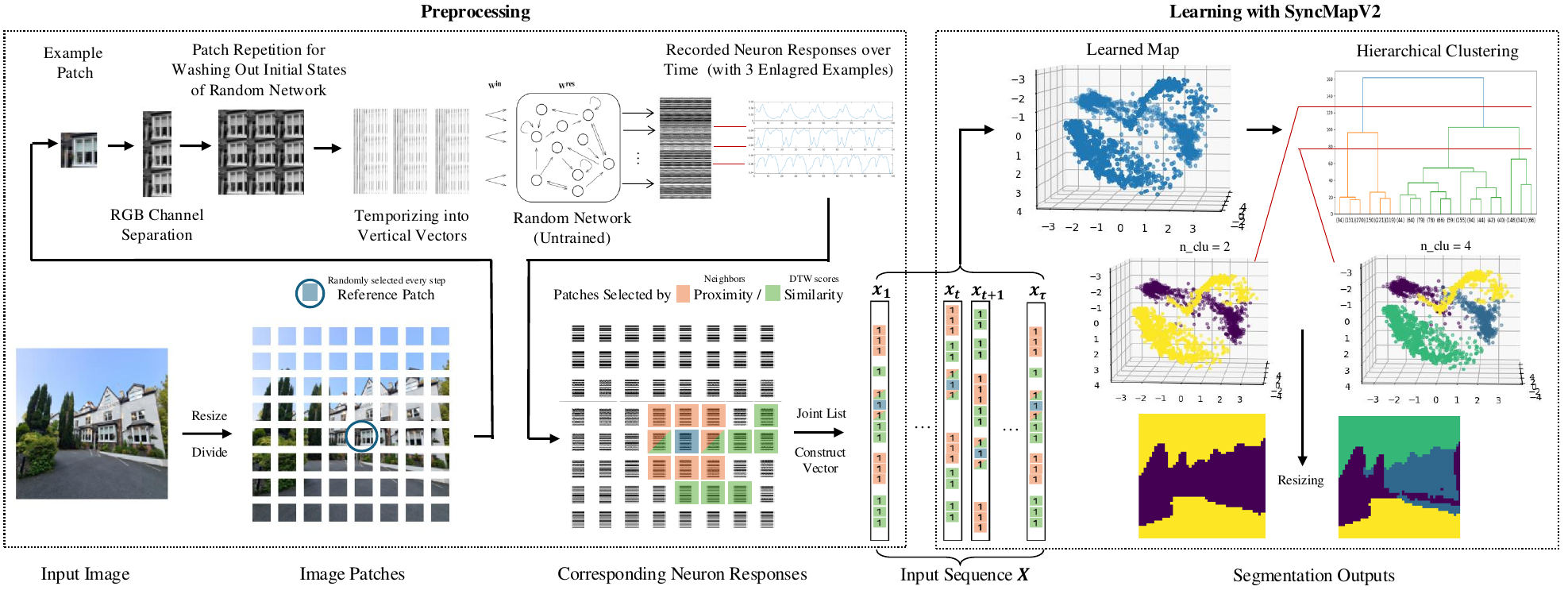} 
% \caption{The workflow be divided into two parts: \textbf{Preprocessing} and \textbf{Learning}. \textbf{Preprocessing:} Images will be split into patches and separated by RGB channels, then be repeated horizontally and temporized vertically into vectors to feed into an untrained random network. The obtained temporal sequences of each patch will be selected to construct a vector as the activate nods in the SyncMapV2 in one timestep by using proximity and similarity rules. \textbf{Learning:} After the dynamical process, a map will be learned by SyncMapV2, and the segments will be obtained by applying hierarchical clustering.}
\caption{Overview of the preprocessing and learning pipeline for unsupervised image segmentation using SyncMapV2. Preprocessing is performed only once, before learning, to generate the input sequence $\boldsymbol{X}$. The term \textit{learning} is used throughout to emphasize that our workflow is without any self-training or robust training, unlike other segmentation tasks such as instance or semantic segmentation \cite{niu2024unsupervised, lan2024smooseg, tian2024diffuse, sick2024unsupervised, hamilton2022unsupervised}.}
\label{fig:methodology_overview}
\end{figure*}

\subsection{SyncMap}
Inspired by neuron group behaviors, SyncMap is a self-organizing nonlinear dynamical system that can learn and adapt to changes in the problem by creating a dynamic map called SyncMap space, which preserves the correlation between state variables \cite{vargas2021syncmap}.
Inside SyncMap space, attractor-repeller pairs are introduced that serve as a set of self-organizing dynamical equations: State variables that attract each other tend to be spatially grouped as clusters, while those with no attraction will be pulled away from each other by repulsion. 

\textbf{Transformations from input to map space.} Mathematically, SyncMap takes a $N$-dimensional temporal sequence as input. 
Consider a matrix-form of an input sequence $\boldsymbol{X}=[\boldsymbol{x}_1,\boldsymbol{x}_2,...,\boldsymbol{x}_t,...,\boldsymbol{x}_{\tau}]$, where $\tau$ is the sequence length.
$\boldsymbol{x}_t={\{x_{1,t},...,x_{N,t}\}}^{T}$ is a vector at time step $t$, and its elements $x_{i,t}$ ($i=1,...,N$), considered as an input node, hold constrain $x_{i,t}\in \{0,1\}:\sum^{N}_{i=1}x_{i,t}=1$ (i.e., each element is either 0 or 1), where $N$ is the total number of input nodes. 

We then generate the same number of $N$ state variables and randomly initialize them in a $k$-dimensional SyncMap space.
Here, the dimensions of $k$ is considered as a neural space that is generated by neuron populations of $k$.
Each state variable represents the combination of $k$-neurons' responses in SyncMap space. 
To combine them together, we denote it by $\boldsymbol{w}_t={\{\vec{w}_{1,t},...,\vec{w}_{N,t}\}}^{T}$, where $\vec{w}_{i,t} \in \mathbb{R}^k$.
With the above knowledge, we define the nonlinear transformations as follows:
\begin{align} \label{eq:trans_x-w}
    % { \scriptstyle 
    \boldsymbol{x} &\mapsto \boldsymbol{w} ,
\end{align}
\begin{align} \label{eq:PS}
    PS_t[\boldsymbol{w}] = \{\vec{w}_{i,t}|x_{i,t}= 1\}, \ NS_t[\boldsymbol{w}] = \{\vec{w}_{i,t}|x_{i,t}= 0\},
\end{align}
or by omitting the term $w$ for short,
\begin{align} \label{eq:NS}
    \ PS_t = \{i|x_{i,t}=1\},  \ \ NS_t = \{i|x_{i,t}= 0\} .
\end{align}

\noindent To illustrate, Eq. \ref{eq:trans_x-w} is an arrow notion that emphasizes the transformation mapping from $\boldsymbol{x}$ in input space to $\boldsymbol{w}$ in SyncMap space. 
Eqs. \ref{eq:PS} and \ref{eq:NS} describe the nonlinear transformations: given an input $\boldsymbol{x}_t$ at time $t$, based on each input node $x_{i,t}$, we divide the corresponding $\vec{w}_{i,t}$ into two sets: 
(1) positive set $PS_t$ including $\vec{w}_{i,t}$ in which its corresponding $x_{i,t}$ are 1, and (2) negative set $NS_t$ including $\vec{w}_{i,t}$ in which its corresponding $x_{i,t}$ are 0. 
% The original SyncMap used $a$ directly at 0.1 \cite{vargas2021syncmap}. 
% In this work, we introduce threshold value $a$, allowing us to achieve more general state memory implementation.

\textbf{Self-organizing Dynamics.} 
SyncMap emphasizes the dynamical features in the space.
This is achieved by creating attractor for $PS_t$ set and repeller for $NS_t$ set.
Specifically, we first calculate the centroids of $PS_t$ and $NS_t$ sets as follows.
At time step $t$, if and only if the cardinality of both sets are greater than one (i.e., \(|PS_t|>1\) and \(|NS_t|>1\)), do:
% \begin{equation} \label{eq:syncmap_cp}
%     % { \scriptstyle 
%     cp_t=\frac{\sum_{i\in PS_t}\vec{w}_{i,t}}{|PS_t|}, \quad
% \end{equation}
% \begin{equation} \label{eq:syncmap_cn}
%     cn_t=\frac{\sum_{i\in NS_t}\vec{w}_{i,t}}{|NS_t|},
%     % }
% \end{equation}
\begin{equation} \label{eq:syncmap_cp}
    % { \scriptstyle 
    cp_t=\frac{\sum_{i\in PS_t}\vec{w}_{i,t}}{|PS_t|}, \quad
% \end{equation}
% \begin{equation} \label{eq:syncmap_cn}
    cn_t=\frac{\sum_{i\in NS_t}\vec{w}_{i,t}}{|NS_t|},
    % }
\end{equation}
where $cp_t$ and $cn_t$ are the centroids of all $\vec{w}_{i}$ in $PS_t$ and all $\vec{w}_{j}$ in $NS_t$ respectively. Notice that when \(|PS_t|\leq 1\) or \(|NS_t|\leq 1\), nothing will be updated in this iteration.
Each state variable (node) $\vec{w}_{i,t}$, corresponding to each input $x_{i,t}$, is updated as follows:
\begin{equation} \label{eq:syncmap_sign}
    % { \scriptstyle 
    \phi_{i,t} =
    \begin{cases} 
     {1, \ i\in PS_t}\\{0, \ i\in NS_t} 
    \end{cases}, 
    \alpha=\begin{cases}
    {\alpha, \ i\in PS_t\cup NS_t}\\{0, \ otherwise}
    \end{cases},
    % }
\end{equation}
 \noindent Positive feedback $F^{+}_{p}\{i\}$:
\begin{equation}
\label{eq:pfb}
    F^{+}_{p}\{i\} = \frac{\phi_{i,t}(cp_t-\vec{w}_{i,t})}{||\vec{w}_{i,t}-cp_t||},
\end{equation}
Negative feedback $F^{-}_{n}\{i\}$: 
\begin{equation}
\label{eq:nfb}
    F^{-}_{n}\{i\} = (-1) * \frac{(1-\phi_{i,t})(cn_t-\vec{w}_{i,t})}{||\vec{w}_{i,t}-cn_t||},
\end{equation}

\begin{equation} \label{eq:syncmap_update}
    \vec{w}_{i,t+1}=\vec{w}_{i,t}+\alpha(F^{+}_{p}\{i\} + F^{-}_{n}\{i\}),
\end{equation}
\noindent where $\alpha$ is the learning rate and $||$·$||$ is the Euclidean distance. All $\vec{w}_{i,t}$ are then normalized to be within a hyper-sphere of $scale$ at each time step. 

\textbf{Clustering.} SyncMap requires a downstream assessment measurement. Given that high correlation between events results in a close and compact distribution in space, a natural solution emerges: to measure these responses, a clustering process is performed after adapting the dynamic to read out the clusters detected by SyncMap. DBSCAN \cite{schubert2017dbscan} and hierarchical clustering \cite{murtagh2012algorithms} are two suggested algorithms used \cite{vargas2021syncmap, zhang2023symmetrical}.

\section{SyncMapV2}
SyncMapV2 introduces several enhancements, making the system more adaptive, faster, stable, and accurate. Details of performance analysis are in Section \ref{sec:ablation_stu}.
\subsection{Adaptive Learning Rate}
In the original work, the learning rate $\alpha$ was held constant throughout the dynamic learning. However, as the system approaches equilibrium, it is desirable for the learning process to decelerate.
One of the significant improvement in this work is to introduce adaptive learning rate for positive and negative feedbacks. 
Specifically, an indicator that the system is approaching equilibrium is when, for most time steps, the average distance of variables in $PS$ to centroid $cp$ is sufficiently small, denoted by $\bar{d}_+$. Under these conditions, the dynamics of the system should be slowed, as it has already adapted well to the input data.
Mathematically, we define $\alpha^+ = \bar{d}_+ / \sqrt{k}$, such that when $\bar{d}_+$ is small, $\alpha^+$ will also be reduced. The update equation is thus modified as follows:
\begin{equation} \label{eq:syncmap_update2}
    \vec{w}_{i,t+1}=\vec{w}_{i,t}+(\alpha^+ *F^{+}_{p}\{i\} + \alpha^- * F^{-}_{n}\{i\}),
\end{equation}
\noindent where $\alpha^-$ is a constant (see Appendix \ref{subsec:ad-lr}).

\subsection{Fast Symmetrical Activations}
We implement symmetrical activations based on the work by \cite{zhang2023symmetrical}, achieving a fivefold increase in processing speed (see Appendix \ref{subsec:syncmap_symm_acti}).

\subsection{Space Normalization}
In the original version, normalization was performed using $\boldsymbol{w}_{norm} = \boldsymbol{w} / \max(\vec{w}_{i,t})$, which introduced discontinuities and instability. We address this by normalizing $\boldsymbol{w}$ to have a mean of 0 and a standard deviation of 1 at each time step.

\subsection{Moving Average}
To provide a more robust evaluation, we introduce a moving average window $L_{\text{movmean}}$, calculating the average over the last $L_{\text{movmean}}$ steps of the map, rather than relying on a single snapshot as in the original work. $L_{\text{movmean}}=2000$ is used in experiment. 

\subsection{Leaking Rate}
We introduce a leaking rate $\beta=0.1$ to modulate the dynamics of SyncMap, providing finer control over the system’s speed.

\section{Unsupervised Segmentation with SyncMapV2}
Gestalt psychologists in the early 20th century developed key theories on how humans perceive visual components as organized wholes rather than isolated parts \cite{wertheimer1938gestalt}. Principles such as similarity, proximity, and good continuation are essential in grouping elements within a visual scene \cite{kandel2000principles}.
Inspired by these concepts, SyncMapV2 uses these perceptual cues (i.e., proximity and similarity) to perform unsupervised image segmentation, organizing visual data into meaningful patterns.
While we are not the first to model human visual system, SyncMapV2 stands out as the first to rely solely on self-organization to achieve robust segmenting performance, even in complex scenarios such as noisy and corrupted environments.

\subsection{Overview}
We begin by dividing the input image into $M_{p} \times N_{p}$ sub-images, or patches, where both $M_{p}$ and $N_{p}$ are set to 48 to achieve an optimal balance between computational efficiency and accuracy. From these patches, we construct an input vector $\boldsymbol{x}_t = \{x_{1,t}, \dots, x_{M_{p} \times N_{p},t}\}^{T}$, where each element $x_{i,t}$ represents the state of the $i$-th patch at time step $t$. If $x_{i,t} = 1$, the patch is selected; if $x_{i,t} = 0$, the patch is not selected at that time step.

Each time step begins by selecting a \textit{reference patch} randomly from all patches. Subsequently, we determine its correlated patches using two key lists: the proximity list $P$ and the similarity list $S$. For the proximity list, the eight neighboring patches surrounding the reference patch are chosen, reflecting the spatial closeness in a 2D image plane. The similarity list is constructed based on a similarity matrix generated by a random network, which measures how closely the features of one patch resemble those of others. See matrix generation process in Section \ref{subsec:usImgSeg_gen_simiMat}.

After selecting the patches, their corresponding elements in the input vector are set to 1 (i.e., $x_{i,t} = 1$ for all selected patches). 
Subsequent steps involve applying the dynamic learning as described in Eqs. 1-\ref{eq:syncmap_update2}. Hierarchical clustering is then applied to effectively group patches into coherent segments or clusters based on their coordinates in SyncMapV2 space, thus producing the final segmented output.

\subsection{Generating Similarity Matrix}
\label{subsec:usImgSeg_gen_simiMat}
Establishing correlations based on similarity between image patches is more complex than proximity-based correlations. Simple methods like mean square error are insufficient due to their sensitivity to noise and high-frequency variations, as well as their focus on pixel values rather than structural similarities. A more robust approach is needed—one that effectively models these similarities by handling the variability and complexity of image data through dynamic patterns rather than static pixel comparisons.

To address this, we apply a random network to transform spatial correlations within the image into temporal sequences. We use an echo state network (ESN) for its simplicity, rich nonlinearity, and low computational cost \cite{jaeger2001echo}. See Appendix \ref{subsec:esn} for details. As a reservoir computing model, ESN focuses less on the details of individual neurons and more on neural population encoding (representational space). This approach aligns better with our research objectives and is generally more robust than a feedforward structure. The workflow for generating neural responses for a particular patch, as outlined in Figure \ref{fig:methodology_overview}, applies this method broadly to all patches.

Initially, the RGB channels of an image patch are separated and stacked vertically to create a grayscale image, which is then repeated $K$ times ($K$=3) to ensure that the initial states of ESN's dynamics are washed out.  This repetition also introduces a periodic pattern in each neuron's response. 
Subsequently, the image is \textit{temporized} by dividing it into 1D vertical vectors (Figure \ref{fig:methodology_overview}(top-left)), each serving as input to the random network. The readout layer of ESN is removed, and we directly use the temporal neural responses from the reservoir neurons. Since each patch is processed using the same random network, their neural responses are comparable. Dynamic Time Warping (DTW) \cite{giorgino2009computing} is then applied to calculate the similarity between patches. See Appendix \ref{subsec:DTW} for details.
Conceptually, each patch generates its own ``similarity matrix,'' recording its similarity to other patches. When a reference patch is selected (Figure \ref{fig:methodology_overview}(bottom-left)), a similarity list ($S$) is created by sorting patches based on their similarity to the reference patch, with the reference patch listed first (zero difference). This list $S$ is then combined with the proximity list ($P$) to construct the input vector $\boldsymbol{x}_t$ for SyncMapV2.

\begin{table}[ht]
\caption{SyncMapV2 outperforms the SOTA methods on the VOC 2012 benchmark (bold) and ranks second on the BSD500 benchmark (underlined). The results are reported as mIoU. The `*’ denotes OIS results. The lower bound represents the mIoU of random segmentation.}
\centering
\resizebox{0.4\columnwidth}{!}{%
\begin{tabular}{lrr}
\toprule
\textbf{Method} & \textbf{VOC2012} & \textbf{BSD500} \\ 
\midrule
Lower bound, n = 2 & 0.2530 & 0.1158 \\
Lower bound, n = 5 & 0.1271 & 0.0683 \\

k-means clustering, k = 2 & 0.3166 & 0.1972 \\
k-means clustering, k = 17 & 0.2383 & 0.2648 \\
IIC, k = 2 & 0.2729 & 0.1733 \\
IIC, k = 20 & 0.2005 & 0.2071 \\
Backprop & 0.3082 & 0.3239 \\
DFC, $\mu$ = 5 & 0.3520 & \textbf{0.3739} \\
Pixel-level* & \underline{0.4103} & - \\
SyncMap* & - & 0.2500 \\
\midrule
\textbf{SyncMapV2}, n = 2 & 0.3900 & 0.2191 \\
\textbf{SyncMapV2}, n = 5 & 0.3539 & 0.2682 \\
\textbf{SyncMapV2*} & \textbf{0.4566} & \underline{0.3279} \\

\bottomrule
\end{tabular}%
}

\label{tab:standard_test}
\end{table}

\section{Experiments and Results}
Recent methods proposed for unsupervised image segmentation have predominantly focused on benchmark tests using clean images \cite{kim2020unsupervised,hoang2024pixel}. However, we argue that these experiments fall short of evaluating a model’s performance in terms of adaptability and robustness, which are crucial for real-world applications where conditions are often unpredictable and data can be irregular. For the first time, we go beyond standard benchmarks by introducing \textbf{robustness tests} in various noisy and corrupted environments, and \textbf{adaptability tests} in continually changing environments.

\subsection{Experimental Setups}
\subsubsection{Dataset} 
We test our model on PASCAL VOC 2012 and BSD500 segmentation benchmarks \cite{everingham2015pascal, arbelaez2010contour}. 
For VOC 2012, we used the ‘`trainval'' set with 2,913 images, treating each segment and the background as individual segments. Since our model is unsupervised, object category labels were ignored.
For BSD500, we standardized the dataset by selecting the annotation with the fewest clusters per image. This aligns with the ``coarse'' setting used in existing works \cite{kim2020unsupervised}, which we also adopt for our terminology.

\subsubsection{Evaluation Metrics}
Following the commonly used methodology \cite{kim2020unsupervised, hoang2024pixel}, we evaluate the segmentation performance using the mean Intersection over Union (mIOU). Here, mIOU was calculated as the mean IOU of each segment in the ground truth (GT) and the estimated segment that had the largest IOU with the GT segment. See Appendix \ref{appendix:measurement} for details.

\subsubsection{Implementation Details}
For SyncMapV2, input images are resized to $288 \times 288$ and divided into $48 \times 48$ patches. We initialize an ESN with 512 neurons, a sparsity of 0.9, and set the input scaling, leaking rate, and spectral radius to 1.0, 0.5, and 1.1, respectively. The SyncMapV2 space dimension is set to $k$=15, with the proximity and similarity lists ($P$ and $S$) limited to the top nine patches each.
After learning for $\tau$=200,000 steps, hierarchical clustering is performed with cluster numbers $n_{clu}$ set from 2 to 20.
The segmentation result is then reshaped back to the original input image size. We report two performance metrics: Optimal Dataset Scale (ODS), which evaluates a fixed $n_{clu}$ across the dataset, and Optimal Image Scale (OIS), which is widely used to select the optimal $n_{clu}$ for each image \cite{hoang2024pixel, zhou2020dic, li2012segmentation}.

\begin{table*}[ht]
\caption{mIoU of robustness tests on BSD500 across four types of corruptions \cite{hendrycks2019robustness}. The best scores are in bold, with percentages indicating the improvements of SyncMapV2 over DFC and SyncMap.}
\resizebox{\textwidth}{!}{%
\begin{tabular}{lc|cccc|cccc|cccc|cccc|c}
\toprule
\textbf{Corruption Type} & \textbf{Clean} & \multicolumn{4}{c|}{\textbf{Gaussian Noise}} & \multicolumn{4}{c|}{\textbf{Zoom Blur}} & \multicolumn{4}{c|}{\textbf{Snow Weather}} & \multicolumn{4}{c|}{\textbf{Digital Contrast}} & \textbf{Overall}\\ 
\midrule
\textbf{Method / Severity} & - & S1 & S3 & S5 & \textbf{Mean} & S1 & S3 & S5 & \textbf{Mean} & S1 & S3 & S5 & \textbf{Mean} & S1 & S3 & S5 & \textbf{Mean} & \textbf{Mean}\\
\midrule
DFC & \textbf{0.37} & 0.26 & 0.23 & 0.21 & 0.23 & 0.27 & 0.26 & 0.26 & 0.26 & 0.26 & 0.25 & 0.23 & 0.25 & 0.27 & 0.27 & 0.25 & 0.26 & 0.26\\
Orig. SyncMap & 0.25 & 0.24 & 0.24 & 0.24 & 0.24 & 0.26 & 0.26 & 0.25 & 0.26 & 0.25 & 0.23 & 0.23 & 0.24 & 0.25 & 0.26 & 0.25 & 0.25 &0.25\\
\textbf{SyncMapV2} & 0.33 & \textbf{0.32} & \textbf{0.31} & \textbf{0.28} & \textbf{0.30} & \textbf{0.32} & \textbf{0.30} & \textbf{0.30} & \textbf{0.31} & \textbf{0.32} & \textbf{0.30} & \textbf{0.29} & \textbf{0.30} & \textbf{0.33} & \textbf{0.33} & \textbf{0.32} & \textbf{0.33} & \textbf{0.31}\\
\midrule
\textit{ (vs. DFC, \%)} & {\color[HTML]{CB0000} -12\%} & {\color[HTML]{009901} 26\%} & {\color[HTML]{009901} 32\%} & {\color[HTML]{009901} 34\%} & {\color[HTML]{009901} \textbf{31\%}} & {\color[HTML]{009901} 16\%} & {\color[HTML]{009901} 16\%} & {\color[HTML]{009901} 16\%} & {\color[HTML]{009901} \textbf{16\%}} & {\color[HTML]{009901} 20\%} & {\color[HTML]{009901} 21\%} & {\color[HTML]{009901} 26\%} & {\color[HTML]{009901} \textbf{23\%}} & {\color[HTML]{009901} 20\%} & {\color[HTML]{009901} 22\%} & {\color[HTML]{009901} 30\%} & {\color[HTML]{009901} \textbf{24\%}} & {\color[HTML]{009901} \textbf{21\%}}\\
\textit{ (vs. Orig. SyncMap, \%)} & {\color[HTML]{009901} 31\%} & {\color[HTML]{009901} 32\%} & {\color[HTML]{009901} 31\%} & {\color[HTML]{009901} 17\%} & {\color[HTML]{009901} \textbf{26\%}} & {\color[HTML]{009901} 21\%} & {\color[HTML]{009901} 17\%} & {\color[HTML]{009901} 18\%} & {\color[HTML]{009901} \textbf{20\%}} & {\color[HTML]{009901} 30\%} & {\color[HTML]{009901} 31\%} & {\color[HTML]{009901} 25\%} & {\color[HTML]{009901} \textbf{27\%}} & {\color[HTML]{009901} 30\%} & {\color[HTML]{009901} 29\%} & {\color[HTML]{009901} 29\%} & {\color[HTML]{009901} \textbf{31\%}} & {\color[HTML]{009901} \textbf{26\%}}\\
\bottomrule
\end{tabular}%
}
\label{tab:robustness_test_rebuttal}
\end{table*}

\begin{figure}[ht]
\centering
\includegraphics[width=0.75\columnwidth]{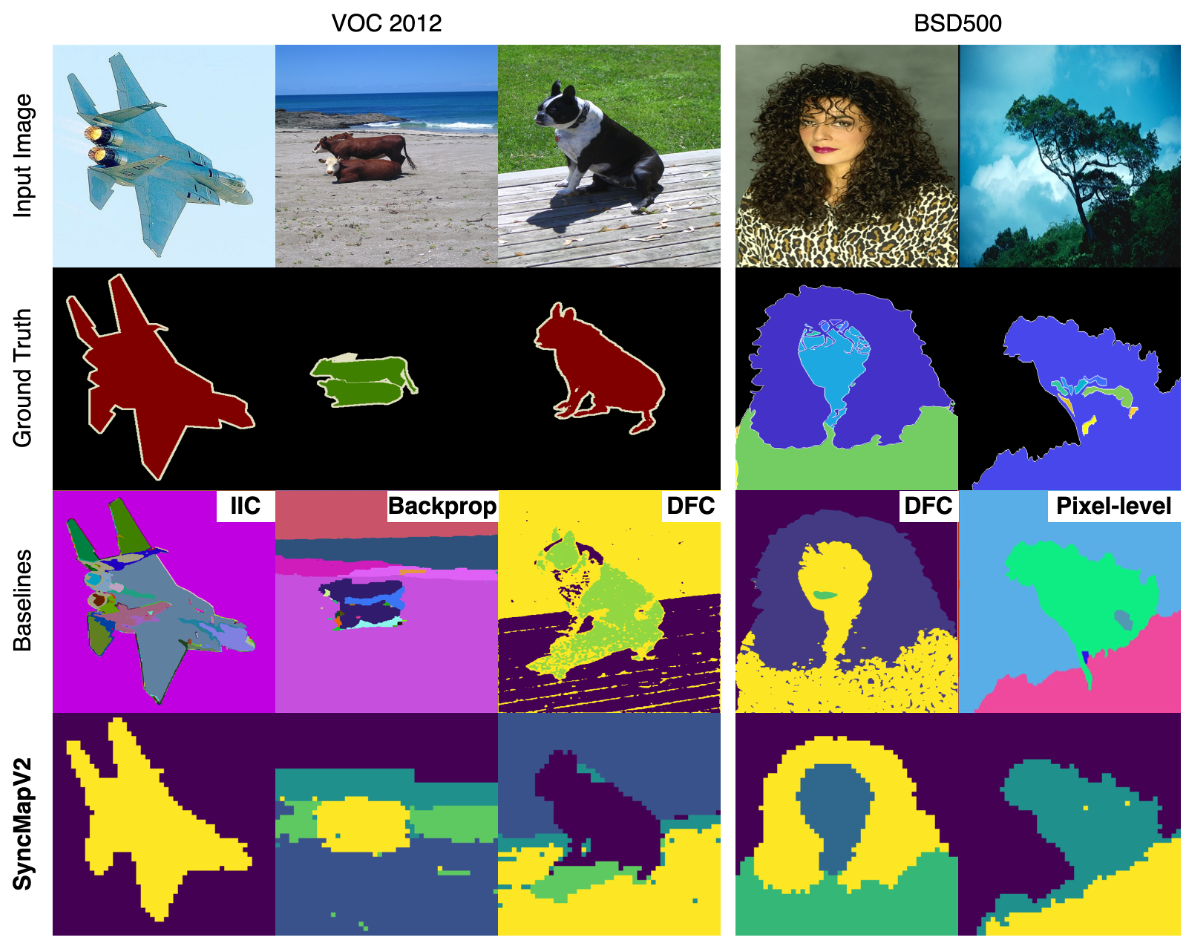} 
\caption{Comparison of segmentation on standard tests. Different segments are shown in different colors.}
\label{fig:standard_comparison}
\end{figure}

\begin{figure}[ht]
\centering
\includegraphics[width=0.75\columnwidth]{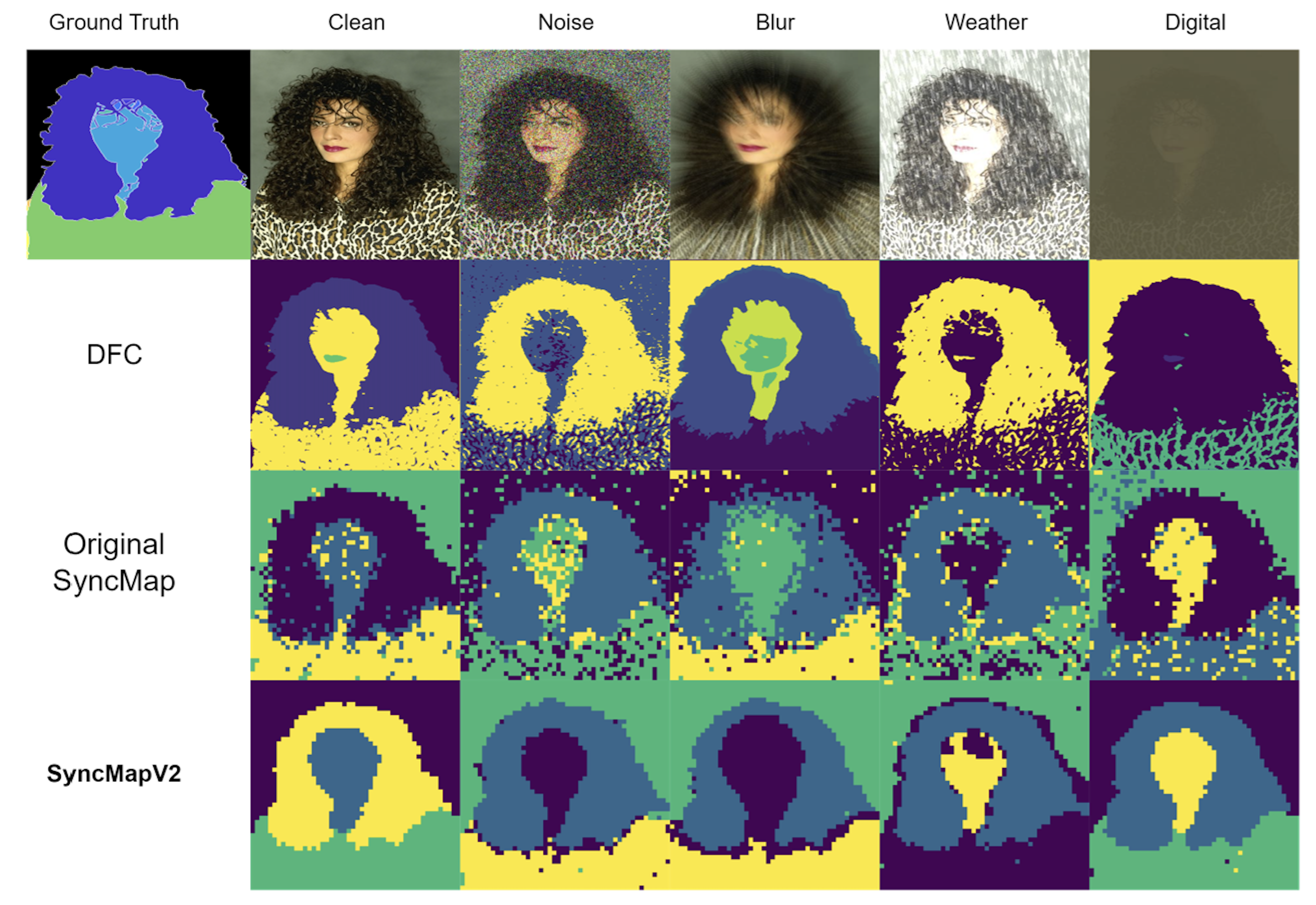} 
% \caption{The comparison between DFC and SyncMapV2 on clean and corrupted images.}
\caption{SyncMapV2 performs robustly across four types of algorithmically generated corruptions from noise, blur, weather, and digital categories. See the settings of severity levels in the Appendix 
\ref{appendix:corruptions}.}
\label{fig:robustness_samples}
\end{figure}

\begin{figure}[ht]
\centering
\includegraphics[width=0.6\columnwidth]{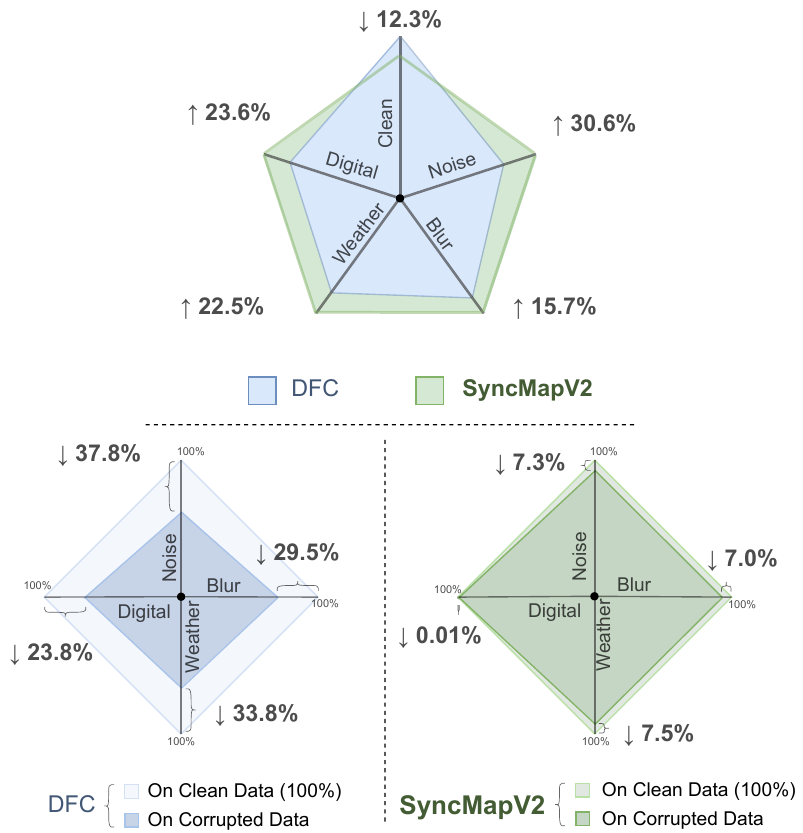} 
\caption{SyncMapV2 outperforms the SOTA method (DFC) across all corrupted conditions (top). Vertically, DFC experiences a 31.2\% performance decrease on average (bottom-left), while SyncMapV2 sees only a 5.5\% drop, with near-zero decrease in digital corruption (bottom-right).}
\label{fig:robustness_comparison}
\end{figure}

\subsubsection{Baseline Settings}
We employ DFC \cite{kim2020unsupervised}, using parameter settings from the original paper and its open-source code. Specifically, we use a maximum of 1000 iterations, a minimum label count of 2, and set the continuity loss $\mu$ to 5, as recommended. Additional baselines include k-means, IIC \cite{Invariant2019iic}, Backprops \cite{Kanezaki2018Unsupervised}, and Pixel-level network \cite{hoang2024pixel}. 

\subsection{Standard Benchmark Tests}
Table \ref{tab:standard_test} shows the performance of SyncMapV2 and other methods on VOC 2012 and BSD500 standard benchmarks. SyncMapV2 achieves the highest mIoU of 0.4566 on VOC 2012, outperforming the SOTA method by 11.3\%. 
On BSD500, SyncMapV2 ranks second, just below the top-performing DFC. The variability in segment numbers within BSD500 makes OIS metric a more effective measure for visualizing the performance of SyncMapV2, as it allows the model to adaptively determine the optimal number of clusters for each image. The lower bounds, which represent the performance of random segmentation, are also included in the table to clarify that the mIoU does not start from zero.

\subsection{Robustness Tests}
To the best of our knowledge, SyncMapV2 represents the first
effort in addressing unsupervised segmentation under corruptions. The evaluation metrics and setup used to assess the robustness of SyncMapV2 and baselines are adopted from established benchmarks \cite{hendrycks2019robustness}. 

The robustness performance is evaluated across four types of corruption, each with increasing levels of severity. Specifically, we used \textit{Gaussian noise}, \textit{zoom blur}, \textit{snow weather}, and \textit{digital contrast} types. The models were tested on corrupted versions of the BSD500 benchmark, with the mean mIoU scores for each type of corruption calculated by averaging the mIoU at severity levels S1, S3, and S5.

As shown in Table \ref{tab:robustness_test_rebuttal} and Figure \ref{fig:robustness_comparison}, while SyncMapV2 performs 12.3\% worse than the SOTA on the clean data, it significantly outperforms on corrupted data, with improvements of 30.6\% under noise, 15.7\% under blur, 22.5\% under weather conditions, and 23.6\% under digital contrast corruption. Notably, SyncMapV2’s mIoU drops only 0.01\% under digital corruption, compared to a substantial 23.8\% drop for the SOTA. This pattern holds across other corruption types as well: for noise, it experiences a 7.3\% decline versus 37.7\% for the SOTA; for weather, a 7.5\% drop versus 33.8\%; and for blur, a 7.0\% drop versus 29.5\%.
\begin{figure}[ht]
\centering
\includegraphics[width=0.75\columnwidth]{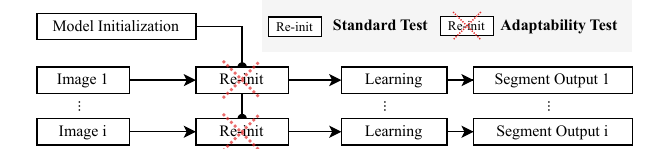}
\caption{We propose the first adaptability test pipeline, where model initialization occurs only once.}
\label{fig:adatability_pipeline}
\end{figure}

These results strongly suggest that while features learned by neural networks with loss functions may perform better on specific clean datasets, they often lack robustness. In contrast, SyncMapV2 distinguishes itself by learning robust features through self-organizing dynamics, making it the first model to achieve levels of resilience in unsupervised segmentation. 
Having said that, achieving this robustness often comes with a trade-off in accuracy on clean data.
To develop models that can reliably operate against adversarial attacks and irregular environments, one must prioritize learning robust features over optimizing for peak performance on standard benchmarks.

\begin{figure}[ht]
\centering
\includegraphics[width=0.6\columnwidth]{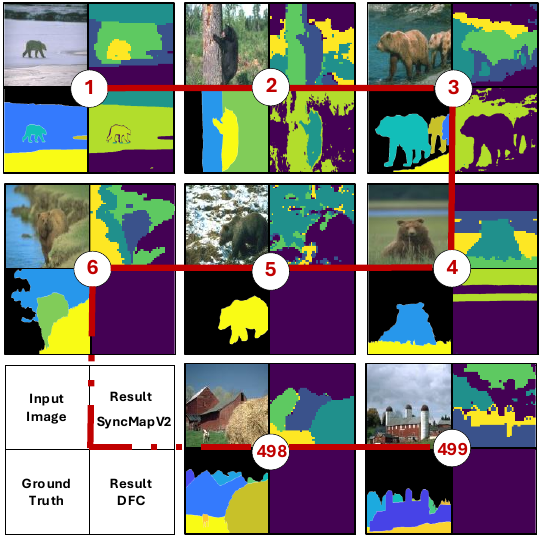}
\caption{Sample results of adaptability tests. SyncMapV2 consistently adapts to each input, while DFC fails to adapt early on (see Figure \ref{fig:loss_plot} where the loss converges by the $5^{th}$ image, hindering further learning).
}
\label{fig:adaptability_samples}
\end{figure}

\begin{figure*}[ht]
\centering
\includegraphics[width=1.0\textwidth]{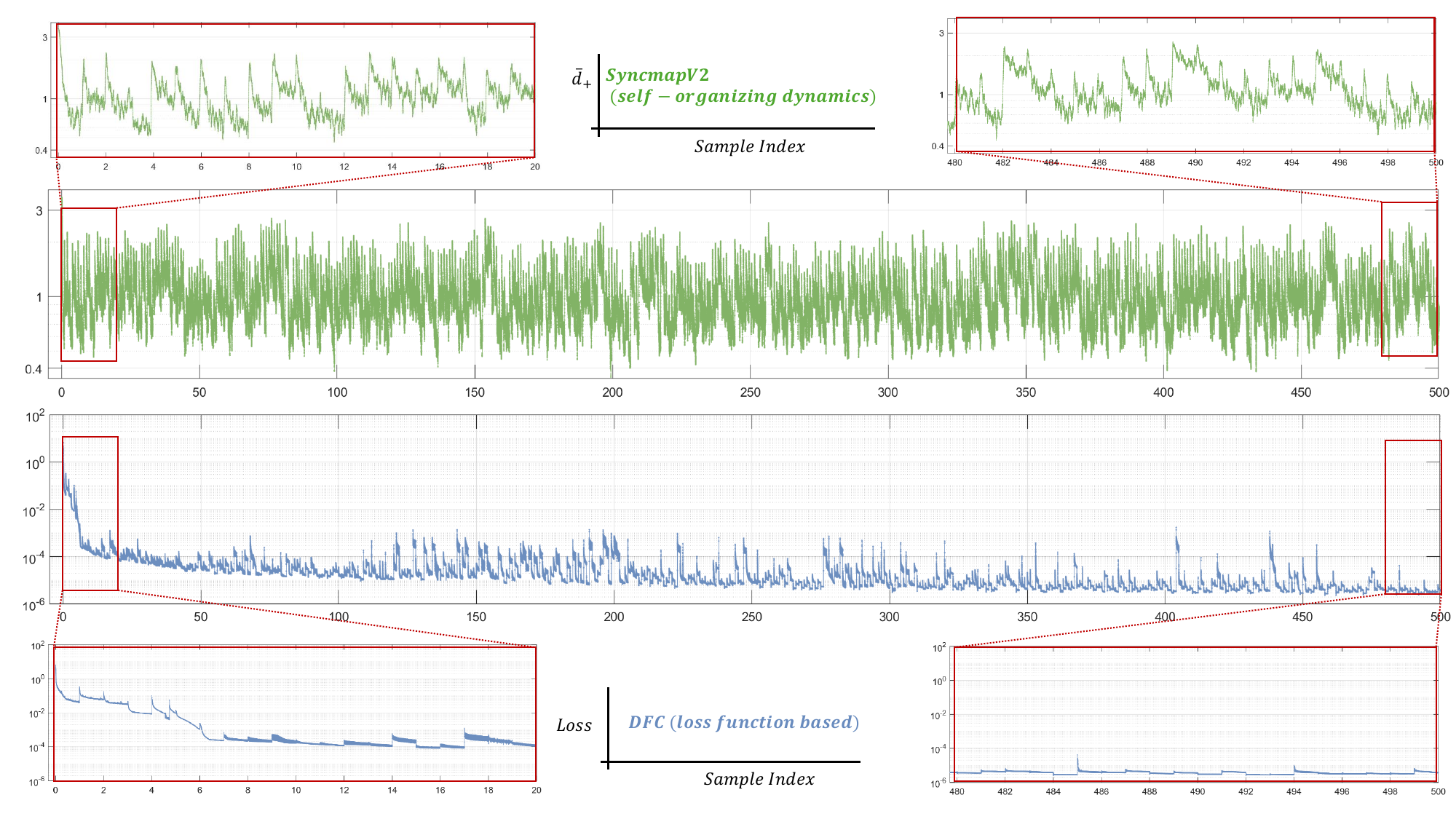} 
\caption{Adaptability plots over time for self-organization-based SyncMapV2 (green, top) and loss function-based DFC (blue, bottom). Both methods were given ample time to learn before the input image was changed. The average distance to the attractor ($\bar{d}_+$) and loss values are recorded across 500 samples, respectively. SyncMapV2 demonstrates inherent adaptability, while DFC fails to adapt (note that the y-axis of the all plots is log-scaled). See Figure \ref{fig:adaptability_samples} for sample results.}
\label{fig:loss_plot}
\end{figure*}

\begin{table}[ht]
\centering

\label{tab:my-table}
\caption{mIoU of adaptability tests on BSD500. The best scores are in bold. ``erl.stp.'' refers to early stopping.}
\resizebox{0.4\columnwidth}{!}{%
\begin{tabular}{lcrr}
\toprule
\textbf{Method} & \textbf{w/ Re-init} & \multicolumn{2}{c}{\textbf{w/o Re-init}} \\ 
\midrule
Lower bound & 0.1642 & 0.1642 & 0.1642 \\ 
\midrule
\begin{tabular}[c]{@{}r@{}} \\ DFC \end{tabular} & \begin{tabular}[c]{@{}r@{}} \\ \textbf{0.3739} \end{tabular} & \begin{tabular}[c]{@{}r@{}}w/ erl.stp. \\ 0.1740 \end{tabular} & \begin{tabular}[c]{@{}r@{}}w/o erl.stp.\\ 0.1657\end{tabular} \\ 
\midrule
\begin{tabular}[c]{@{}r@{}} \\ \textbf{SyncMapV2} \end{tabular} & \begin{tabular}[c]{@{}r@{}} \\ 0.3279 \end{tabular} & \begin{tabular}[c]{@{}r@{}}$\tau$ \\ \textbf{0.3227}\end{tabular} & \begin{tabular}[c]{@{}r@{}}$2\tau$ \\ \textbf{0.3292}\end{tabular} \\ 
\bottomrule
\end{tabular}%
}
\label{tab:adaptability_test}
\end{table}

\subsection{Adaptability Tests}
Standard procedures for unsupervised image segmentation include model re-initialization for each new input image. This is a common practice in most loss function-based methods \cite{Kanezaki2018Unsupervised,Invariant2019iic,kim2020unsupervised,hoang2024pixel,guermazi2024dynaseg}. In contrast, human vision operates continuously, adapting seamlessly to new scenes.
Here, we present SyncMapV2 as the first model capable of adapting to new inputs with near-zero performance degradation, without the need for re-initialization.
We conducted extensive experiments on the BSD500 dataset, initializing each model only once at the start of the learning process  (Figure \ref{fig:adatability_pipeline}). To further quantify adaptability, we extended the sequence length for SyncMapV2 to $2\tau$=400,000. For DFC, we tested two settings: (1) early stopping upon reaching a minimum label count, and (2) without early stopping, allowing learning to continue to the maximum iterations.

Results in Table \ref{tab:adaptability_test} clearly demonstrate SyncMapV2’s superior adaptability compared to loss functions-based method. 
% SyncMapV2 exhibits a near-zero performance decrease when operating without re-initialization. 
No statistically significant difference is observed for SyncMapV2, with mIoU slightly drops to 0.3227 ($\tau$) and improves to 0.3292 ($2\tau$). 
% This near-zero performance decrease underscores SyncMapV2’s ability to continuously adapt to changes in input images without significant loss in segmentation accuracy.
This adaptability is expected given SyncMapV2’s design, which steadily updates the correlations between state variables (i.e., corresponding to each image patch) projected into the map space. 
As observed in Figure \ref{fig:loss_plot}, given an input image, the average distance $\bar{d}_+$ decreases gradually at the beginning of learning and stabilizes once the dynamics reach equilibrium. However, the change of input images affects the place of attractors, naturally causing the system to temporarily enter an unstable state and initiating the adaptation. This is indicated by the increase and subsequent stabilization of $\bar{d}_+$ over time.

In stark contrast, the DFC method exhibits a huge drop in performance without re-initialization, with mIoU of 0.1740 and 0.1657 with and without early stopping, respectively—values only slightly above the lower bound of 0.1642. This suggests that DFC, similar to other loss function-based methods, struggles to learn when the loss converges after processing a few images. The model’s learning capability is severely hindered, resulting in outputs that are practically meaningless, as illustrated in Figure \ref{fig:adaptability_samples}. This lack of adaptability highlights a critical limitation of optimization-based approaches in dynamic environments.

\begin{table*}[ht]
\caption{Ablation Study. Green big-O represent technique applied.}
\resizebox{\textwidth}{!}{%
\begin{tabular}{l|ccccc|cc|cc}
\toprule
\textbf{} & \multicolumn{5}{c|}{\textbf{SyncMapV2 Ablation}} & \multicolumn{2}{c|}{\textbf{mIoU}} & \multicolumn{2}{c}{\textbf{Imporvements from SyncMap}}\\ 
\midrule
\textbf{Model} & Adaptive Learning Rate & Symmetrical Activations & Space Normalization & Moving Average & Leaking Rate & Clean & Gaussian S3 & Clean & Gaussian S3\\
\midrule
Orig. SyncMap & - & - & - & - & - & 0.251 & 0.236 & 0\% & 0\%\\
\midrule
AdpLR & {\color[HTML]{009901} \textbf{O}} & - & - & - & - & 0.290 & 0.255 & 15.5\% & 8.1\%\\
SymmAct & - & {\color[HTML]{009901} \textbf{O}} & - & - & - & 0.162 & 0.143 & -35.5\% & -39.5\%\\
SNorm & - & - & {\color[HTML]{009901} \textbf{O}} & - & - & 0.282 & 0.258 & 12.4\% & 9.3\%\\
MovAvg & - & - & - & {\color[HTML]{009901} \textbf{O}} & - & 0.301 & 0.282 & 19.9\% & 19.5\%\\
LeakR & - & - & - & - & {\color[HTML]{009901} \textbf{O}} & 0.282 & 0.255 & 12.4\% & 8.1\%\\
SNorm + LeakR & - & - & {\color[HTML]{009901} \textbf{O}} & - & {\color[HTML]{009901} \textbf{O}} & 0.31 & 0.28 & 23.0\% & 19.1\%\\
SymmAct + AdpLR & {\color[HTML]{009901} \textbf{O}} & {\color[HTML]{009901} \textbf{O}} & - & - & - & 0.17 & 0.20 & -33.6\% & -17.0\%\\
SymmAct + SNorm & - & {\color[HTML]{009901} \textbf{O}} & {\color[HTML]{009901} \textbf{O}} & - & - & 0.30 & 0.27 & 18.9\% & 12.9\%\\
SymmAct + MovAvg & - & {\color[HTML]{009901} \textbf{O}} & - & {\color[HTML]{009901} \textbf{O}} & - & 0.14 & 0.15 & -45.2\% & -36.8\%\\
SymmAct + LeakR & - & {\color[HTML]{009901} \textbf{O}} & - & - & {\color[HTML]{009901} \textbf{O}} & 0.28 & 0.27 & 10.5\% & 14.5\%\\
SymmAct + SNorm + LeakR & - & {\color[HTML]{009901} \textbf{O}} & {\color[HTML]{009901} \textbf{O}} & - & {\color[HTML]{009901} \textbf{O}} & 0.32 & 0.29 & 28.4\% & 22.4\%\\
\midrule
SyncMapV2 & {\color[HTML]{009901} \textbf{O}} & {\color[HTML]{009901} \textbf{O}} & {\color[HTML]{009901} \textbf{O}} & {\color[HTML]{009901} \textbf{O}} & {\color[HTML]{009901} \textbf{O}} & {\color[HTML]{009901} \textbf{0.33}} & {\color[HTML]{009901} \textbf{0.31}} & {\color[HTML]{009901} \textbf{31.5\%}} & {\color[HTML]{009901} \textbf{31.4\%}}\\
\bottomrule
\end{tabular}%
}
\label{tab:ablation_test_rebuttal}
\end{table*}

\subsection{Ablation Study}
\label{sec:ablation_stu}
Table \ref{tab:ablation_test_rebuttal} shows an ablation study on various configurations of the SyncMapV2 model, highlighting the contributions of individual components. 
Each model variation tests a different combination of settings, including Adaptive Learning Rate (AdpLR), Symmetrical Activations (SymmAct), Space Normalization (SNorm), Moving Average (MovAvg), and Leaking Rate (LeakR). 
The table reports the Mean Intersection over Union (mIoU) scores on both clean data and data with Gaussian noise (S3), as well as the improvement percentages over the original SyncMap model.

\subsubsection{Adaptive Learning Rate}
AdpLR dynamically adjusts the learning rate based on system dynamics, specifically aiding in the control of fast-moving dynamics as the system approaches equilibrium. This helps stabilize the dynamics, leading to an improvement of 15.5\% under clean conditions and 8.1\% under noisy conditions. The results show that AdpLR reduces over-responsiveness and improves the accuracy of convergence, allowing for a more robust segmentation process.

\subsubsection{Symmetrical Activation}
While SymmAct was originally proposed in prior work to balance dynamics effectively in a temporal setting, it did not perform as expected when applied independently to SyncMap for image segmentation. In isolation, it caused a performance drop (-35.5\% in clean, -39.5\% under noise). However, when combined with other techniques, SymmAct contributes to a balanced update of dynamics, especially when paired with stabilizing mechanisms like AdpLR and MovAvg. This finding highlights SymmAct's role as a complementary, rather than standalone, component in this application.

\subsubsection{Space Normalization}
SNorm standardizes the spatial distribution to zero mean and unit variance at each time step, addressing potential skew in spatial points and improving clustering consistency. By providing a stable spatial reference, SNorm offers a 12.4\% improvement in clean conditions and 9.3\% under noise. This component ensures that patch similarity is consistently evaluated, which aids in accurate grouping, especially under noisy conditions.

\subsubsection{Moving Average}
MovAvg provides temporal smoothing by averaging the system's dynamics over a defined history window, thus capturing broader patterns in patch similarity rather than relying on a single-time snapshot. MovAvg shows the most significant improvement on its own, with a gain of 19.9\% in clean conditions and 19.5\% under noise, underscoring its importance in achieving stable segmentation results. By reducing “snapshot bias,” MovAvg enables more reliable clustering in both clean and noisy environments.

\subsubsection{Leaking Rate}
LeakR introduces a global rate to moderate the speed of system dynamics, preventing the system from fluctuating too rapidly. This slower dynamic allows the system to “settle” more predictably, supporting stable clustering and segmentation. LeakR achieves a performance gain of 12.4\% in clean conditions and 8.1\% under noise, showing its impact in aligning the dynamics to a steady progression as equilibrium approaches.

\subsubsection{Combined System (SyncMapV2)}
Integrating all five components achieves the highest performance boost, with an improvement of 31.5\% in clean and 31.4\% in noisy conditions. This synergy shows that each component supports and complements the others:
\begin{itemize}
    \item AdpLR, MovAvg, and LeakR stabilize and smooth the dynamics, facilitating controlled and reliable convergence.
    \item SNorm ensures spatial consistency for clustering, making group formations more stable and accurate.
    \item SymmAct acts as a balance-enhancing factor when combined with the other components.
\end{itemize}

\subsubsection{Adaptability Test}
In addition to robustness test, we also applied the original SyncMap \cite{vargas2021syncmap} and SyncMapV2 (this work) on the BSD500 benchmark under adaptability test settings. The results show that SyncMapV2 achieved an mIoU of 0.3292, outperforming the original SyncMap, which scored 0.2975. 
This enhancement underscores the effectiveness of the improvements incorporated into SyncMapV2.

\section{Conclusion}
We present SyncMapV2, an improved self-organizing dynamical system that models the robustness and adaptability of human vision in unsupervised segmentation tasks—capabilities that existing systems often lack. SyncMapV2 demonstrates state-of-the-art resilience for the first time, maintaining high performance across noisy and corrupted data. It is also the first to achieve near-zero performance decrease with dynamically changing inputs, without the need for re-initialization. SyncMapV2 sets a new benchmark in unsupervised segmentation, paving the way for future research in systems that require robust and adaptive learning in real-world environments.

\appendix 
\section{Measurement Metric}
\label{appendix:measurement}
\subsection{Mean Intersection over Union (mIoU)}

Following the commonly used methodology \cite{kim2020unsupervised, hoang2024pixel}, we evaluate the segmentation performance using the mean Intersection over Union (mIoU). Here, mIoU was calculated as the mean IoU of each segment in the ground truth (GT) and the estimated segment that had the largest IoU with the GT segment. The IoU for a particular segment \(s\) in the ground truth is calculated as:

\[
\text{IoU}(s) = \frac{|P_s \cap G_s|}{|P_s \cup G_s|}
\]
where \(P_s\) represents the set of predicted pixels for segment \(s\), and \(G_s\) represents the set of ground truth pixels for segment \(s\). The mIoU is then computed as:

\[
\text{mIoU} = \frac{1}{S} \sum_{s=1}^{S} \max_{p \in P} \text{IoU}(s, p)
\]
where \(S\) is the total number of ground truth segments, and \(P\) is the set of predicted segments. The final mIoU score is the average of the highest IoU scores between each ground truth segment and the predicted segments.

\subsection{Unsupervised mIoU Calculation Implementation}

The following Python code demonstrates the implementation of the unsupervised mIoU metric. The code first preprocesses the predicted and ground truth labels, then calculates the IoU for each pair of segments, selecting the maximum IoU for each ground truth segment.

\begin{algorithm}
\caption{Unsupervised IoU Calculation}
\begin{algorithmic}
\Function{unsupervised\_iou}{prediction, label\_name, img\_W, img\_H}
    \State label\_map = label\_preprocess(label\_name)
    \State temp\_img = Image.fromarray(prediction)
    \State temp\_img = temp\_img.resize((img\_W, img\_H), resample=Image.Resampling.BOX)
    \State prediction = np.asarray(temp\_img)
    \State prediction = np.reshape(prediction, [img\_H * img\_W])
    \State predict\_map = prediction\_preprocess(prediction)

    \State metric = tf.keras.metrics.IoU(num\_classes=2, target\_class\_ids=[1])
    \State total\_scores = []

    \For{\textbf{each} label in label\_map}
        \State score = []
        \For{\textbf{each} predict in predict\_map}
            \State binary\_metric = tf.keras.metrics.BinaryIoU(target\_class\_ids=[1])
            \State binary\_metric.update\_state(label, predict)
            \State score.append(binary\_metric.result().numpy())
        \EndFor
        \State total\_scores.append(max(score))
    \EndFor

    \State \Return total\_scores
\EndFunction

\Function{unsupervised\_meaniou}{prediction, label\_name, img\_W, img\_H}
    \State mean\_iou = np.array(unsupervised\_iou(prediction, label\_name, img\_W, img\_H)).mean()
    \State \Return mean\_iou
\EndFunction
\end{algorithmic}
\end{algorithm}
This code snippet demonstrates how to calculate the unsupervised mIoU by considering the maximum IoU for each ground truth segment across the predicted segments.

\section{Methodology}
\label{appendix:method}
\subsection{Random Networks}
\label{subsec:esn}
\begin{figure}[ht]
    \centering
    \includegraphics[width=0.5\columnwidth]{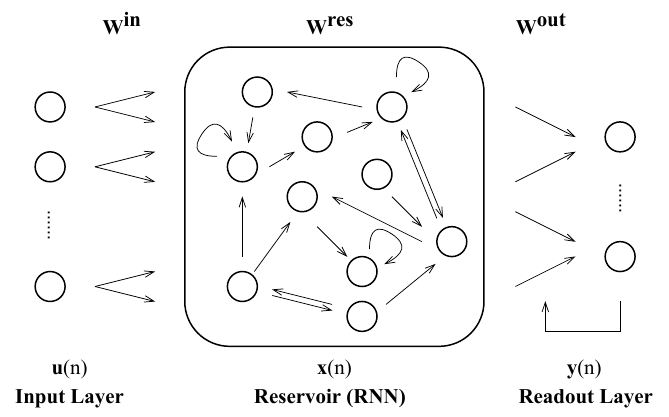}
    \caption[Simplified structure of Echo State Network]{Simplified structure of Echo State Network (ESN).
    }
    \label{fig:ESN archi}
\end{figure} 

We applied an echo state network (ESN) as a random network to transform the spatial information to temporal information. ESN was first proposed by \cite{jaeger2001echo}. 
To illustrate the technical details, here we use the notations by \cite{lukovsevivcius2012practical}. Note that we omit the readout layer $\mathbf{W}^{out}$, since we directly use the temporal responses generated by the neurons.
Consider a temporal processing task, where the input signal is $\mathbf{u}(n) \in \mathbb{R}^{N_{u}}$, given $n=1,...,T$ with $T$ being the total number of discrete data points. 
The simplified update equations of the reservoir part in ESNs are given by:

\begin{equation}
    \label{eq:esn1}
    \mathbf{\Tilde{x}}(n)=tanh(\mathbf{W}^{in}\mathbf{u}(n)+
    \mathbf{W}\mathbf{x}(n-1)),
\end{equation}
\begin{equation}
    \label{eq:esn2}
    \mathbf{x}(n)=(1-\alpha)\mathbf{x}(n-1) + \alpha\mathbf{\Tilde{x}}(n),
\end{equation}

\noindent where 
$\mathbf{\Tilde{x}}(n) \in \mathbb{R}^{N_{x}}$ is the update at time step $n$,
$\mathbf{x}(n) \in \mathbb{R}^{N_{x}}$ is the state vector of the reservoir neurons (also known as the resulting states or the \textit{echo} of its input history \cite{lukovsevivcius2009reservoir}), 
$\mathbf{W}^{in} \in \mathbb{R}^{N_{x} \times N_{u}}$ and $\mathbf{W} \in \mathbb{R}^{N_{x} \times N_{x}}$ are the weight matrices of the input-reservoir connections and the recurrent connections inside the reservoir, respectively. $tanh()$ is the non-linear activation function applied element-wise. $\alpha$ is the leaking rate that mainly controls the speed of the dynamics. 

We recorded every $\mathbf{x}(n), n=1,...,T$ for every image patch. After that, we perform Dynamic Time Warping (DTW) to calculate the ``patch-wise'' similarity using $tslearn$ Python package. 

\noindent \textbf{Key parameters to define the random network:}
\begin{quote}
    
\begin{enumerate}
    \item \textbf{Population size ($\mathbf{W}$):}
    \begin{itemize}
        \item Defines how many neurons are in the reservoir (i.e., the shape of $\mathbf{W}$), we set it as 512, therefore the shape of weight matrix $\mathbf{W}$ is $(512,512)$.
    \end{itemize}

    \item \textbf{Spectral radius (\(\rho\)):}
    \begin{itemize}
        \item The largest absolute eigenvalue of the reservoir's weight matrix. It controls the stability of the dynamics in the reservoir. We set $\rho=1.1$.
    \end{itemize}

    \item \textbf{Input size and matrix ($\mathbf{W}^{in}$):}
    \begin{itemize}
        \item Determines the shape of the input matrix $\mathbf{W}^{in}$. Let the size of resized images be $N_r\times N_r$ and the number of blocks that an image will be split as $M\times M$, We set this as $\frac {N_r}{M} \times 3$, in practice, $N_r=288$ and $M=48$, therefore $\text{Input Size} =\left\lfloor \frac{288}{48} \right\rfloor \times 3=18$, so the shape of $\mathbf{W}^{in}$ is $(512, 18)$
    \end{itemize}

    \item \textbf{Sparsity:}
    \begin{itemize}
        \item The proportion of non-zero connections in the reservoir's weight matrix. Sparse connectivity is often used to create more diverse and dynamic states in the reservoir. We set $\text{Sparsity}=0.9$.
    \end{itemize}

    \item \textbf{Leaking rate ($\alpha$):}
    \begin{itemize}
        \item A parameter that controls the update speed of the reservoir states. It balances the influence of past states versus new inputs. We set $\alpha=0.5$.
    \end{itemize}

    \item \textbf{Reservoir input and weight matrix ($\mathbf{W}^{in}$\&$\mathbf{W}$) Initialization:}
    \begin{itemize}
        \item The method used to initialize the weights of the connections between reservoir neurons. Common methods include random initialization from a uniform or normal distribution, often scaled by the spectral radius. In practice, we apply normal distribution.
    \end{itemize}

    \item \textbf{Input and weight scaling:}
    \begin{itemize}
        \item Determines the strength of the input signal when fed into the reservoir. It's important for controlling how much influence the input has on the reservoir's dynamics. In practice, we set input scaling as $1.0$ and weight scaling as $2.0$.
    \end{itemize}
\end{enumerate}

\end{quote}

\subsection{Dynamic Time Warping}
\label{subsec:DTW}

Dynamic Time Warping (DTW) \cite{giorgino2009computing} is a similarity measure between time series. Let us consider two time series \( \mathbf{x} = \{x_1, x_2, \dots, x_n\} \) and \( \mathbf{y} = \{y_1, y_2, \dots, y_m\} \) of respective lengths \( n \) and \( m \). Here, all elements \( x_i \) and \( y_j \) are assumed to lie in the same \( d \)-dimensional space. In \texttt{tslearn}, such time series would be represented as arrays of respective shapes \((n, d)\) and \((m, d)\), and DTW can be computed using the following code:

\begin{quote}
    
\begin{verbatim}
from tslearn.metrics import dtw, dtw_path

dtw_score = dtw(x, y)
\end{verbatim}
\end{quote}

\subsection{Symmetrical Activation}
\label{subsec:syncmap_symm_acti}
We applied symmetrical activation \cite{zhang2023symmetrical}, where equal number of positive and negative nodes are selected to activate at each time step.

\subsubsection{Stochastically Select Nodes into $\boldsymbol{NS_t}$ Set}
\label{subsubsec:stochastic_select_ns}
After obtaining the $PS_t$, we define the temporary negative set $NS_{temp}$=$W_t$\ -\ $PS_t$, where $W_t=\{w_{i,t}|i=1,...,n\}$ is the set including all nodes.

Next, we use the stochastic selection for sampling several negative nodes in $NS_{temp}$ set. The number of negative nodes being selected is symmetrically equal to the cardinality of $PS_t$ (i.e., $|NS_t|=|PS_t|$). 
After this step, the $NS_t$ set is updated ($NS_{t} \subseteq NS_{temp}$), which only includes the activated negative nodes.

The remaining steps follow the Equations 4-9 in main text. We calculate a moving average of $L_{mean}$ steps of nodes' position and use it for clustering, instead of applying clustering to the map at a ``snapshot'' time step in the original work.

\subsection{Adaptive Learning Rate in SyncMapV2}
\label{subsec:ad-lr}

In the main text, we introduced an adaptive learning rate strategy to improve the learning dynamics as the system approaches equilibrium. This approach is particularly important because, as the system stabilizes, continuing with a high learning rate might lead to unnecessary oscillations or instability. The adaptive learning rate allows the system to decelerate its learning as it gets closer to equilibrium, ensuring more precise and stable convergence.

\subsubsection{Mathematical Formulation}

As described in the main text, the adaptive learning rate for the positive feedback loop, denoted as \( \alpha^+ \), is given by:

\[
\alpha^+ = \frac{\bar{d}_+}{\sqrt{k}},
\]

where \( \bar{d}_+ \) is the average distance of variables in the positive set \( PS \) to the centroid \( cp \), and \( k \) represents the dimensionality of the space. The corresponding update equation is:

\begin{equation} \label{eq:syncmap_update2_}
    \vec{w}_{i,t+1}=\vec{w}_{i,t}+(\alpha^+ \cdot F^{+}_{p}\{i\} + \alpha^- \cdot F^{-}_{n}\{i\}),
\end{equation}

Here, \( \alpha^- \) remains constant, but \( \alpha^+ \) adapts based on the system's proximity to equilibrium.

\subsubsection{Implementation Details}

The following algorithmic description demonstrates how the adaptive learning rate is computed and updated during the learning process:

\begin{algorithm}
\caption{Update Adaptive Learning Rate}
\begin{algorithmic}
\Function{update\_adaptive\_learning\_rate}{dist\_set2centroid \_positive, set\_positive}
    \State dist\_avg\_positive = np.sum(dist\_set2centroid\_positive \(\times\) set\_positive[:, None]) / set\_positive.sum()
    \State adaptive\_LR\_positive = dist\_avg\_positive / self.space\_scale\_dimensions\_sqrt

    % Save history for analysis
    \If{np.random.rand() smaller than self.variable\_history\_record\_freq}
        \State self.variable\_history.append(dist\_avg\_positive)
    \EndIf

    % Update the learning rate using Widrow-Hoff rule
    \State self.adaptive\_LR += self.adaptive\_LR\_WH \(\times\) (adaptive\_LR\_positive - self.adaptive\_LR)
    \State adaptive\_LR\_positive = copy.deepcopy(self.adaptive\_LR)
    \State adaptive\_LR\_negative = 1

    % Adjustments for negative feedback
    \State adaptive\_LR\_negative\_amplifier\_a = 0.01
    \State adaptive\_LR\_negative\_amplifier\_b = 2
    \State adaptive\_LR\_negative\_amplifier = self.input\_size \(\times\) adaptive\_LR\_negative\_amplifier\_a + adaptive\_LR\_negative\_amplifier\_b
    \State adaptive\_LR\_negative\_threshold = 1.5
    \State adaptive\_LR\_negative = self.adaptive\_LR \(\times\) adaptive\_LR\_negative\_amplifier
    \If{adaptive\_LR\_negative greater than adaptive\_LR\_negative\_threshold}
        \State adaptive\_LR\_negative = adaptive\_LR\_negative\_threshold
    \EndIf

    % Thresholding the adaptive learning rate for stability
    \State adaptive\_LR\_positive\_threshold = 0.05
    \If{adaptive\_LR\_positive small than adaptive\_LR\_positive\_threshold}
        \State \Return adaptive\_LR\_positive\_threshold, adaptive\_LR\_negative
    \Else
        \State \Return adaptive\_LR\_positive, adaptive\_LR\_negative
    \EndIf
\EndFunction
\end{algorithmic}
\end{algorithm}

\subsubsection{Explanation of the Implementation}

The code begins by calculating the average distance of variables in the positive set to the centroid, denoted as \( \texttt{dist\_avg\_positive} \). This distance is then used to compute the initial adaptive learning rate for positive feedback, \( \texttt{adaptive\_LR\_positive} \), which is scaled by the dimensionality of the space (\( \sqrt{k} \) in the equation, equivalent to \texttt{self.space\_scale\_dimensions\_sqrt} in the code).

The learning rate for negative feedback, \( \alpha^- \), remains constant but is amplified according to the system's input size and additional scaling factors. The final learning rates are thresholded to prevent instability, ensuring \( \alpha^+ \) does not become too small or too large, which would otherwise affect the system's ability to learn effectively.

This adaptive learning rate mechanism thus ensures that as the system approaches equilibrium, the positive feedback learning rate \( \alpha^+ \) is reduced, allowing the system to fine-tune its adaptation without overshooting or causing oscillations.

\begin{table*}[ht]
\caption{Statistical test between different models or settings.}
\centering
\resizebox{0.7\textwidth}{!}{%
\begin{tabular}{llll}
\toprule
\textbf{Task}       & \textbf{Description}              & \textbf{Test}                & \textbf{P-Value ($\alpha=0.05$)}                                                                   \\ \midrule
Noise               & SyncMapV2 vs. DFC                 & Independent Two-Sided T-Test & \begin{tabular}[c]{@{}l@{}}S1: 3.28E-12\\ S3: 2.27E-16\\ S5: 1.34E-16\end{tabular} \\
Weather             & SyncMapV2 vs. DFC                 & Independent Two-Sided T-Test & \begin{tabular}[c]{@{}l@{}}S1: 3.28E-08\\ S3: 3.65E-08\\ S5: 6.55E-11\end{tabular} \\
Blur                & SyncMapV2 vs. DFC                 & Independent Two-Sided T-Test & \begin{tabular}[c]{@{}l@{}}S1: 5.77E-06\\ S3: 5.57E-06\\ S5: 5.16E-06\end{tabular} \\
Digital             & SyncMapV2 vs. DFC                 & Independent Two-Sided T-Test & \begin{tabular}[c]{@{}l@{}}S1: 4.27E-07\\ S3: 2.94E-09\\ S5: 2.20E-14\end{tabular} \\ \midrule
No re-initialization & SyncMapV2 vs. DFC                 & Independent Two-Sided T-Test & 3.93E-63                                                                           \\
Varied sequence length                & SyncMapV2 $\tau$ vs. SyncMapV2 $2\tau$ & Paired Two-Sided T-Test      & 3.46E-01                                                                           \\ \bottomrule
\end{tabular}
}
\end{table*}

% \usepackage{graphicx}
\begin{table*}[ht]
\caption{Paired Two-Sided T-Test of SyncMapV2 P-value. Avg: Average mIoU of severity 1, 3, and 5. We compare the average performance of each corruption type with performance on clean images.}
\centering
\resizebox{0.6\textwidth}{!}{%
\begin{tabular}{llllll}
\toprule
\textbf{Compared with Clean} & \textbf{Outinit} & \textbf{Noise Avg} & \textbf{Blur Avg} & \textbf{Weather Avg} & \textbf{Digital Avg} \\
\midrule
n=2                          & 6.99E-01         & 2.06E-01           & 1.81E-04          & 3.27E-01             & 2.82E-01             \\
n=3                          & 6.89E-01         & 4.58E-05           & 1.95E-09          & 1.13E-02             & 1.81E-03             \\
n=4                          & 4.69E-01         & 1.68E-05           & 2.68E-09          & 9.00E-03             & 6.89E-01             \\
n=5                          & 1.86E-01         & 1.00E-07           & 9.60E-17          & 3.35E-03             & 3.58E-01             \\
n=6                          & 9.94E-02         & 6.63E-08           & 3.29E-18          & 7.04E-04             & 4.81E-01             \\
n=7                          & 9.76E-03         & 2.39E-07           & 5.87E-17          & 6.95E-04             & 9.66E-01             \\
n=8                          & 3.24E-03         & 3.61E-05           & 5.93E-19          & 1.31E-03             & 7.02E-01             \\
n=9                          & 1.40E-03         & 6.40E-05           & 7.10E-23          & 9.32E-04             & 5.73E-01             \\
n=10                         & 7.38E-04         & 2.44E-04           & 6.43E-25          & 2.41E-03             & 2.07E-01             \\
n=11                         & 1.50E-03         & 3.00E-04           & 5.79E-27          & 2.48E-03             & 2.76E-01             \\
n=12                         & 8.08E-04         & 1.82E-04           & 4.72E-30          & 1.23E-03             & 3.50E-01             \\
n=13                         & 5.07E-04         & 5.71E-04           & 2.07E-32          & 1.89E-03             & 3.50E-01             \\
n=14                         & 8.78E-03         & 1.43E-04           & 3.16E-30          & 6.37E-04             & 9.35E-01             \\
n=15                         & 5.05E-03         & 9.03E-05           & 5.54E-31          & 4.65E-04             & 8.74E-01             \\
n=16                         & 6.54E-03         & 3.68E-05           & 3.09E-32          & 2.68E-04             & 6.60E-01             \\
n=17                         & 2.05E-03         & 1.29E-04           & 1.11E-30          & 5.79E-04             & 8.10E-01             \\
n=18                         & 2.15E-03         & 1.94E-04           & 5.45E-31          & 1.13E-03             & 5.50E-01             \\
n=19                         & 1.03E-03         & 2.91E-04           & 9.92E-31          & 1.65E-03             & 3.91E-01             \\
n=20                         & 5.93E-03         & 5.38E-04           & 2.33E-30          & 2.36E-03             & 5.46E-01             \\
% n=30                         & 1.09E-01         & 4.89E-07           & 1.29E-37          & 5.66E-05             & 9.90E-01             \\
OIS                          & 2.40E-02         & 4.57E-27           & 2.91E-26          & 1.18E-09             & 1.96E-01            \\
\bottomrule
\end{tabular}%
}
\end{table*}

\section{Corruptions and Severity Level}
\label{appendix:corruptions}
In this paper, we applied image corruption to BSDS500 images, to accomplish it, we used \textbf{ImageCorruptions} library.

The \textbf{ImageCorruptions} project provides a Python package for corrupting images, designed to test the robustness of neural networks against various perturbations. It includes common corruptions like Gaussian blur, noise, and more, with different severity levels. The package supports both RGB and grayscale images of any size and aspect ratio and is available for installation via pip. This tool is primarily for benchmarking rather than data augmentation.
For more details, visit the https://github.com/bethgelab/imagecorruptions. 
% And for the segmented samples of corrupted images, please check the \textbf{Segmented Images} file.

% \lipsum
\section{Statistical Analysis}
\label{appendix:stat}
To assess the performance differences between models or settings, we conducted statistical tests using p-values at a significance level of $\alpha = 0.05$. The results are summarized in the table below.

\begin{itemize}
    \item \textbf{Task}: The specific aspect of the model being evaluated, such as Noise, Weather, Blur, or Digital.
    \item \textbf{Test}: The statistical test used for the comparison. Most comparisons were conducted using an Independent Two-Sided T-Test, which checks if there is a significant difference between the means of two independent groups. In some cases, a Paired Two-Sided T-Test was used, which is appropriate when the two sets of data are related or paired.
\end{itemize}

\bibliographystyle{unsrt}  
\bibliography{main}

\begin{thebibliography}{10}

\bibitem{azulay2019deep}
Aharon Azulay and Yair Weiss.
\newblock Why do deep convolutional networks generalize so poorly to small image transformations?
\newblock {\em Journal of Machine Learning Research}, 20(184):1--25, 2019.

\bibitem{recht2018cifar}
Benjamin Recht, Rebecca Roelofs, Ludwig Schmidt, and Vaishaal Shankar.
\newblock Do cifar-10 classifiers generalize to cifar-10?
\newblock {\em arXiv preprint arXiv:1806.00451}, 2018.

\bibitem{cea2014intriguing}
Szegedy Cea.
\newblock Intriguing properties of neural networks.
\newblock {\em In ICLR. Citeseer}, 2014.

\bibitem{goodfellow2014explaining}
Ian~J Goodfellow, Jonathon Shlens, and Christian Szegedy.
\newblock Explaining and harnessing adversarial examples.
\newblock {\em arXiv preprint arXiv:1412.6572}, 2014.

\bibitem{nguyen2015deep}
Anh Nguyen, Jason Yosinski, and Jeff Clune.
\newblock Deep neural networks are easily fooled: High confidence predictions for unrecognizable images.
\newblock In {\em Proceedings of the IEEE conference on computer vision and pattern recognition}, pages 427--436, 2015.

\bibitem{kotyan2022adversarial}
Shashank Kotyan and Danilo~Vasconcellos Vargas.
\newblock Adversarial robustness assessment: Why in evaluation both {L$_0$} and {L$_\infty$} attacks are necessary.
\newblock {\em Plos one}, 17(4):e0265723, 2022.

\bibitem{buckman2018thermometer}
Jacob Buckman, Aurko Roy, Colin Raffel, and Ian Goodfellow.
\newblock Thermometer encoding: One hot way to resist adversarial examples.
\newblock In {\em International conference on learning representations}, 2018.

\bibitem{uesato2018adversarial}
Jonathan Uesato, Brendan O’donoghue, Pushmeet Kohli, and Aaron Oord.
\newblock Adversarial risk and the dangers of evaluating against weak attacks.
\newblock In {\em International Conference on Machine Learning}, pages 5025--5034. PMLR, 2018.

\bibitem{madry2017towards}
Aleksander Madry, Aleksandar Makelov, Ludwig Schmidt, Dimitris Tsipras, and Adrian Vladu.
\newblock Towards deep learning models resistant to adversarial attacks.
\newblock {\em arXiv preprint arXiv:1706.06083}, 2017.

\bibitem{xu2017feature}
Weilin Xu, David Evans, and Yanjun Qi.
\newblock Feature squeezing: Detecting adversarial examples in deep neural networks.
\newblock {\em arXiv preprint arXiv:1704.01155}, 2017.

\bibitem{carlini2017adversarial}
Nicholas Carlini and David Wagner.
\newblock Adversarial examples are not easily detected: Bypassing ten detection methods.
\newblock In {\em Proceedings of the 10th ACM workshop on artificial intelligence and security}, pages 3--14, 2017.

\bibitem{vargas2021syncmap}
Danilo~Vasconcellos Vargas and Toshitake Asabuki.
\newblock Continual general chunking problem and syncmap.
\newblock {\em Proceedings of the AAAI Conference on Artificial Intelligence}, 35(11):10006--10014, 5 2021.

\bibitem{niu2024unsupervised}
Dantong Niu, Xudong Wang, Xinyang Han, Long Lian, Roei Herzig, and Trevor Darrell.
\newblock Unsupervised universal image segmentation.
\newblock In {\em Proceedings of the IEEE/CVF Conference on Computer Vision and Pattern Recognition}, pages 22744--22754, 2024.

\bibitem{hamilton2022unsupervised}
Mark Hamilton, Zhoutong Zhang, Bharath Hariharan, Noah Snavely, and William~T Freeman.
\newblock Unsupervised semantic segmentation by distilling feature correspondences.
\newblock {\em arXiv preprint arXiv:2203.08414}, 2022.

\bibitem{1000236}
D.~Comaniciu and P.~Meer.
\newblock Mean shift: a robust approach toward feature space analysis.
\newblock {\em IEEE Transactions on Pattern Analysis and Machine Intelligence}, 24(5):603--619, 2002.

\bibitem{felzenszwalb2004efficient}
Pedro~F Felzenszwalb and Daniel~P Huttenlocher.
\newblock Efficient graph-based image segmentation.
\newblock {\em International journal of computer vision}, 59:167--181, 2004.

\bibitem{xia2017w}
Xide Xia and Brian Kulis.
\newblock W-net: A deep model for fully unsupervised image segmentation.
\newblock {\em arXiv preprint arXiv:1711.08506}, 2017.

\bibitem{Kanezaki2018Unsupervised}
Asako Kanezaki.
\newblock Unsupervised image segmentation by backpropagation.
\newblock In {\em 2018 IEEE International Conference on Acoustics, Speech and Signal Processing (ICASSP)}, pages 1543--1547, 2018.

\bibitem{Invariant2019iic}
Xu~Ji, Andrea Vedaldi, and Joao Henriques.
\newblock Invariant information clustering for unsupervised image classification and segmentation.
\newblock In {\em 2019 IEEE/CVF International Conference on Computer Vision (ICCV)}, pages 9864--9873, 2019.

\bibitem{kim2020unsupervised}
Wonjik Kim, Asako Kanezaki, and Masayuki Tanaka.
\newblock Unsupervised learning of image segmentation based on differentiable feature clustering.
\newblock {\em IEEE Transactions on Image Processing}, 29:8055--8068, 2020.

\bibitem{hoang2024pixel}
Cuong~Manh Hoang and Byeongkeun Kang.
\newblock Pixel-level clustering network for unsupervised image segmentation.
\newblock {\em Engineering Applications of Artificial Intelligence}, 127:107327, 2024.

\bibitem{barack2021two}
David~L Barack and John~W Krakauer.
\newblock Two views on the cognitive brain.
\newblock {\em Nature Reviews Neuroscience}, 22(6):359--371, 2021.

\bibitem{trautmann2019accurate}
Eric~M Trautmann, Sergey~D Stavisky, Subhaneil Lahiri, Katherine~C Ames, Matthew~T Kaufman, Daniel~J O’Shea, Saurabh Vyas, Xulu Sun, Stephen~I Ryu, Surya Ganguli, et~al.
\newblock Accurate estimation of neural population dynamics without spike sorting.
\newblock {\em Neuron}, 103(2):292--308, 2019.

\bibitem{rigotti2013importance}
Mattia Rigotti, Omri Barak, Melissa~R Warden, Xiao-Jing Wang, Nathaniel~D Daw, Earl~K Miller, and Stefano Fusi.
\newblock The importance of mixed selectivity in complex cognitive tasks.
\newblock {\em Nature}, 497(7451):585--590, 2013.

\bibitem{zhang2023survey}
Heng Zhang and Danilo~Vasconcellos Vargas.
\newblock A survey on reservoir computing and its interdisciplinary applications beyond traditional machine learning.
\newblock {\em IEEE Access}, 2023.

\bibitem{lan2024smooseg}
Mengcheng Lan, Xinjiang Wang, Yiping Ke, Jiaxing Xu, Litong Feng, and Wayne Zhang.
\newblock Smooseg: smoothness prior for unsupervised semantic segmentation.
\newblock {\em Advances in Neural Information Processing Systems}, 36, 2024.

\bibitem{tian2024diffuse}
Junjiao Tian, Lavisha Aggarwal, Andrea Colaco, Zsolt Kira, and Mar Gonzalez-Franco.
\newblock Diffuse attend and segment: Unsupervised zero-shot segmentation using stable diffusion.
\newblock In {\em Proceedings of the IEEE/CVF Conference on Computer Vision and Pattern Recognition}, pages 3554--3563, 2024.

\bibitem{sick2024unsupervised}
Leon Sick, Dominik Engel, Pedro Hermosilla, and Timo Ropinski.
\newblock Unsupervised semantic segmentation through depth-guided feature correlation and sampling.
\newblock In {\em Proceedings of the IEEE/CVF Conference on Computer Vision and Pattern Recognition}, pages 3637--3646, 2024.

\bibitem{schubert2017dbscan}
Erich Schubert, J{\"o}rg Sander, Martin Ester, Hans~Peter Kriegel, and Xiaowei Xu.
\newblock Dbscan revisited, revisited: why and how you should (still) use dbscan.
\newblock {\em ACM Transactions on Database Systems (TODS)}, 42(3):1--21, 2017.

\bibitem{murtagh2012algorithms}
Fionn Murtagh and Pedro Contreras.
\newblock Algorithms for hierarchical clustering: an overview.
\newblock {\em Wiley Interdisciplinary Reviews: Data Mining and Knowledge Discovery}, 2(1):86--97, 2012.

\bibitem{zhang2023symmetrical}
Heng Zhang and Danilo~Vasconcellos Vargas.
\newblock Symmetrical syncmap for imbalanced general chunking problems.
\newblock {\em Physica D: Nonlinear Phenomena}, 456:133923, 2023.

\bibitem{wertheimer1938gestalt}
Max Wertheimer.
\newblock Gestalt psychology.
\newblock {\em Source Book of Gestalt Psychology. New York: Harcourt, Brace and Co}, 1938.

\bibitem{kandel2000principles}
Eric~R Kandel, James~H Schwartz, Thomas~M Jessell, Steven Siegelbaum, A~James Hudspeth, Sarah Mack, et~al.
\newblock {\em Principles of neural science}, volume~4.
\newblock McGraw-hill New York, 2000.

\bibitem{jaeger2001echo}
Herbert Jaeger.
\newblock The “echo state” approach to analysing and training recurrent neural networks-with an erratum note.
\newblock {\em Bonn, Germany: German National Research Center for Information Technology GMD Technical Report}, 148(34):13, 2001.

\bibitem{giorgino2009computing}
Toni Giorgino.
\newblock Computing and visualizing dynamic time warping alignments in r: the dtw package.
\newblock {\em Journal of statistical Software}, 31:1--24, 2009.

\bibitem{everingham2015pascal}
Mark Everingham, SM~Ali Eslami, Luc Van~Gool, Christopher~KI Williams, John Winn, and Andrew Zisserman.
\newblock The pascal visual object classes challenge: A retrospective.
\newblock {\em International journal of computer vision}, 111:98--136, 2015.

\bibitem{arbelaez2010contour}
Pablo Arbelaez, Michael Maire, Charless Fowlkes, and Jitendra Malik.
\newblock Contour detection and hierarchical image segmentation.
\newblock {\em IEEE transactions on pattern analysis and machine intelligence}, 33(5):898--916, 2010.

\bibitem{zhou2020dic}
Lei Zhou and Weiyufeng Wei.
\newblock Dic: deep image clustering for unsupervised image segmentation.
\newblock {\em Ieee Access}, 8:34481--34491, 2020.

\bibitem{li2012segmentation}
Zhenguo Li, Xiao-Ming Wu, and Shih-Fu Chang.
\newblock Segmentation using superpixels: A bipartite graph partitioning approach.
\newblock In {\em 2012 IEEE conference on computer vision and pattern recognition}, pages 789--796. IEEE, 2012.

\bibitem{hendrycks2019robustness}
Dan Hendrycks and Thomas Dietterich.
\newblock Benchmarking neural network robustness to common corruptions and perturbations.
\newblock {\em Proceedings of the International Conference on Learning Representations}, 2019.

\bibitem{guermazi2024dynaseg}
Boujemaa Guermazi, Riadh Ksantini, and Naimul Khan.
\newblock Dynaseg: A deep dynamic fusion method for unsupervised image segmentation incorporating feature similarity and spatial continuity.
\newblock {\em Image and Vision Computing}, page 105206, 2024.

\bibitem{lukovsevivcius2012practical}
Mantas Luko{\v{s}}evi{\v{c}}ius.
\newblock A practical guide to applying echo state networks.
\newblock In {\em Neural networks: Tricks of the trade}, pages 659--686. Springer, 2012.

\bibitem{lukovsevivcius2009reservoir}
Mantas Luko{\v{s}}evi{\v{c}}ius and Herbert Jaeger.
\newblock Reservoir computing approaches to recurrent neural network training.
\newblock {\em Computer Science Review}, 3(3):127--149, 2009.

\end{thebibliography}

\end{document}